\documentclass[letterpaper]{article} 
\usepackage{aaai2026}  
\usepackage{times}  
\usepackage{helvet}  
\usepackage{courier}  
\usepackage[hyphens]{url}  
\usepackage{graphicx} 
\usepackage{natbib}  
\usepackage{caption} 
\usepackage{algorithm}
\usepackage{algorithmic}

\usepackage{newfloat}
\usepackage{listings}
\DeclareCaptionStyle{ruled}{labelfont=normalfont,labelsep=colon,strut=off} 
\floatstyle{ruled}
\newfloat{listing}{tb}{lst}{}
\floatname{listing}{Listing}
\title{HiFi-Mesh: High-Fidelity Efficient 3D Mesh Generation via Compact Autoregressive Dependence}
\author{
    Yanfeng Li\textsuperscript{\rm 1},
    Tao Tan\textsuperscript{\rm 1},
    Qinquan Gao\textsuperscript{\rm 2,\rm 6},
    Zhiwen Cao\textsuperscript{\rm 3},
    Xiaohong Liu\textsuperscript{\rm 4,5}\thanks{Corresponding author},
    Yue Sun\textsuperscript{\rm 1}\footnotemark[1]
}
\affiliations{
	\begingroup\setlength{\parindent}{0pt}
	\textsuperscript{\rm 1}Intelligent Medical Computing Laboratory, Faculty of Applied Sciences, Macao Polytechnic University\\
	\textsuperscript{\rm 2}Fuzhou University \\
	\textsuperscript{\rm 3}Sichuan University \\
	\textsuperscript{\rm 4}Shanghai Jiao Tong University \\
	\textsuperscript{\rm 5}Shanghai Jiao Tong University Sichuan Research Institute \\
	\textsuperscript{\rm 6}Imperial Vision Technology Co. Ltd \\
	yuesun@mpu.edu.mo, xiaohongliu@sjtu.edu.cn
	\endgroup
}

\usepackage{bibentry}
\usepackage{amsmath}
\usepackage{amssymb}
\usepackage{pifont}
\usepackage{xcolor}
\definecolor{blockgray}{RGB}{156, 163, 175}
\definecolor{blockK}{RGB}{169, 169, 169}
\usepackage[percent]{overpic}
\usepackage{multirow}
\usepackage{adjustbox}
\usepackage{soul}
\usepackage{arydshln}
\usepackage{graphicx}
\usepackage{booktabs}
\usepackage{float}
\definecolor{tablegreen}{RGB}{218, 232, 252}
\definecolor{tableblue}{RGB}{242, 183, 173} 
\definecolor{tablegray}{RGB}{218, 232, 252}

\begin{document}

\maketitle

\begin{abstract}
High-fidelity 3D meshes can be tokenized into one-dimension (1D) sequences and directly modeled using autoregressive approaches for faces and vertices. However, existing methods suffer from insufficient resource utilization, resulting in slow inference and the ability to handle only small-scale sequences, which severely constrains the expressible structural details.
We introduce the Latent Autoregressive Network (LANE), which incorporates compact autoregressive dependencies in the generation process, achieving a $6\times$ improvement in maximum generatable sequence length compared to existing methods. 
To further accelerate inference, we propose the Adaptive Computation Graph Reconfiguration (AdaGraph) strategy, which effectively overcomes the efficiency bottleneck of traditional serial inference through spatiotemporal decoupling in the generation process.
Experimental validation demonstrates that LANE achieves superior performance across generation speed, structural detail, and geometric consistency, providing an effective solution for high-quality 3D mesh generation.
\end{abstract}


\section{1\ \ \ \ Introduction}
Automatic generation of high-quality 3D meshes is a core task in 3D vision \cite{sinha2025marvel}. With the rapid development of virtual reality, digital twins, and metaverse applications, there is growing demand for large-scale, accurate, and topologically 3D content generation \cite{pan2025vasa, melnik2025digital, kurai2025magiccraft}. This makes automatic 3D mesh generation technology practically significant.

In recent years, Large Reconstruction Models \cite{hong2023lrm} have enabled large-scale 3D reconstruction by predicting Gaussian splatting. However, the resulting point cloud representations lack topological connectivity, limiting their use in downstream applications. Researchers have thus extended this paradigm to generate 3D meshes with topological structure \cite{xu2024instantmesh, wu2024unique3d, li2025craftsman3d}. Yet Gaussian splatting-based mesh generation suffers from coarse geometric details and poor topological accuracy due to fundamental differences between volumetric representations and discrete mesh structures, causing precision loss during conversion. Subsequent research has shifted toward directly learning Signed Distance Field (SDF) mesh representations \cite{zhang2024clay, lai2025hunyuan3d, wu2025direct3d}. However, SDF methods still lose discrete face index information, resulting in poor topological structure with artifacts such as over-smoothing and surface irregularities.
\begin{figure}[!t]
	\centering
	\begin{overpic}[width=0.48\textwidth, trim=0 35 0 0, clip]{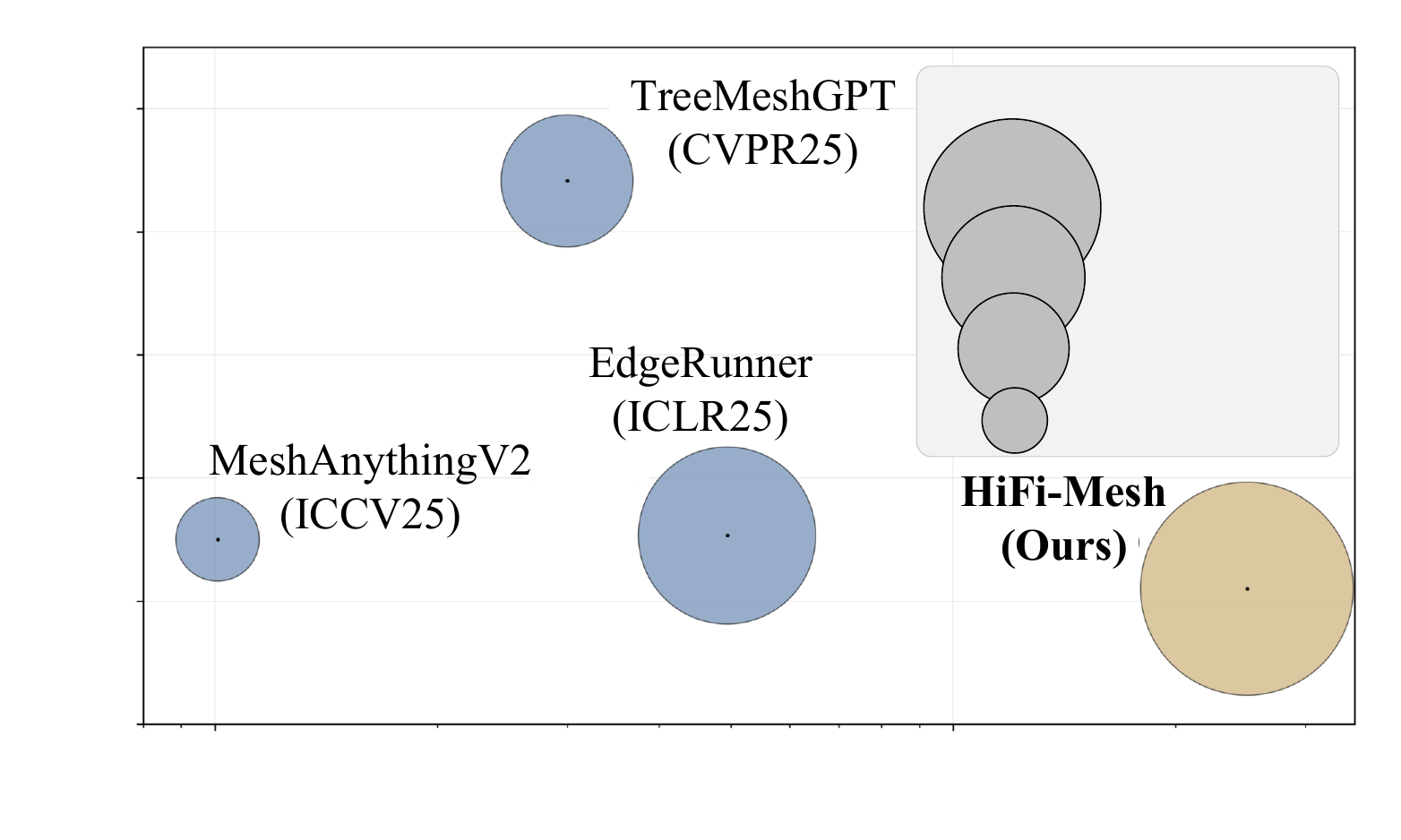}
	\put(4,12.5){\scriptsize\parbox{0.5cm}{\centering{{0.0}}}}
	\put(4,4.5){\scriptsize\parbox{0.5cm}{\centering{{-0.1}}}}
	\put(4,21.5){\scriptsize\parbox{0.5cm}{\centering{{0.1}}}}
	\put(4,30.5){\scriptsize\parbox{0.5cm}{\centering{{0.2}}}}
	\put(4,39.5){\scriptsize\parbox{0.5cm}{\centering{{0.3}}}}
	\put(4,48.5){\scriptsize\parbox{0.5cm}{\centering{{0.4}}}}
	\put(2.25,26.7){\rotatebox[origin=c]{90}{\scriptsize\parbox{2.5cm}{\centering{{Chamfer Distance$\downarrow$}}}}}
	\put(13,1){\scriptsize\parbox{0.5cm}{\centering{{$10^4$}}}}
	\put(67,1){\scriptsize\parbox{0.5cm}{\centering{{$10^5$}}}}
	\put(22,-2){\scriptsize\parbox{5.5cm}{\centering{{Maximum Generation Sequence Length$\uparrow$}}}}
	\put(74,47){\scriptsize\parbox{2cm}{\centering{{Inference Speed$\uparrow$}}}}
	\put(76.5,41.5){\scriptsize\parbox{2cm}{\centering{{302.1 Tok/s}}}}
	\put(76.5,35.5){\scriptsize\parbox{2cm}{\centering{{78.4 Tok/s}}}}
	\put(76.5,30){\scriptsize\parbox{2cm}{\centering{{67.3 Tok/s}}}}
	\put(76.5,25){\scriptsize\parbox{2cm}{\centering{{52.8 Tok/s}}}}
	\end{overpic}
	\caption{Performance comparisons of EdgeRunner \cite{tang2024edgerunner}, MeshAnythingV2 \cite{chen2024meshanything}, TreeMeshGPT \cite{lionar2025treemeshgpt}, and HiFi-Mesh (Ours) on Chamferi Distance $\downarrow$, Inference Speed $\uparrow$, and Maximum Generatable Sequence Length $\uparrow$.}
	\vspace{-0pt}
	\label{fig:introduction_display}
\end{figure}
\begin{figure*}[!t]
	\centering
	\begin{overpic}[width=1.0\textwidth, trim=0 20 0 0, clip]{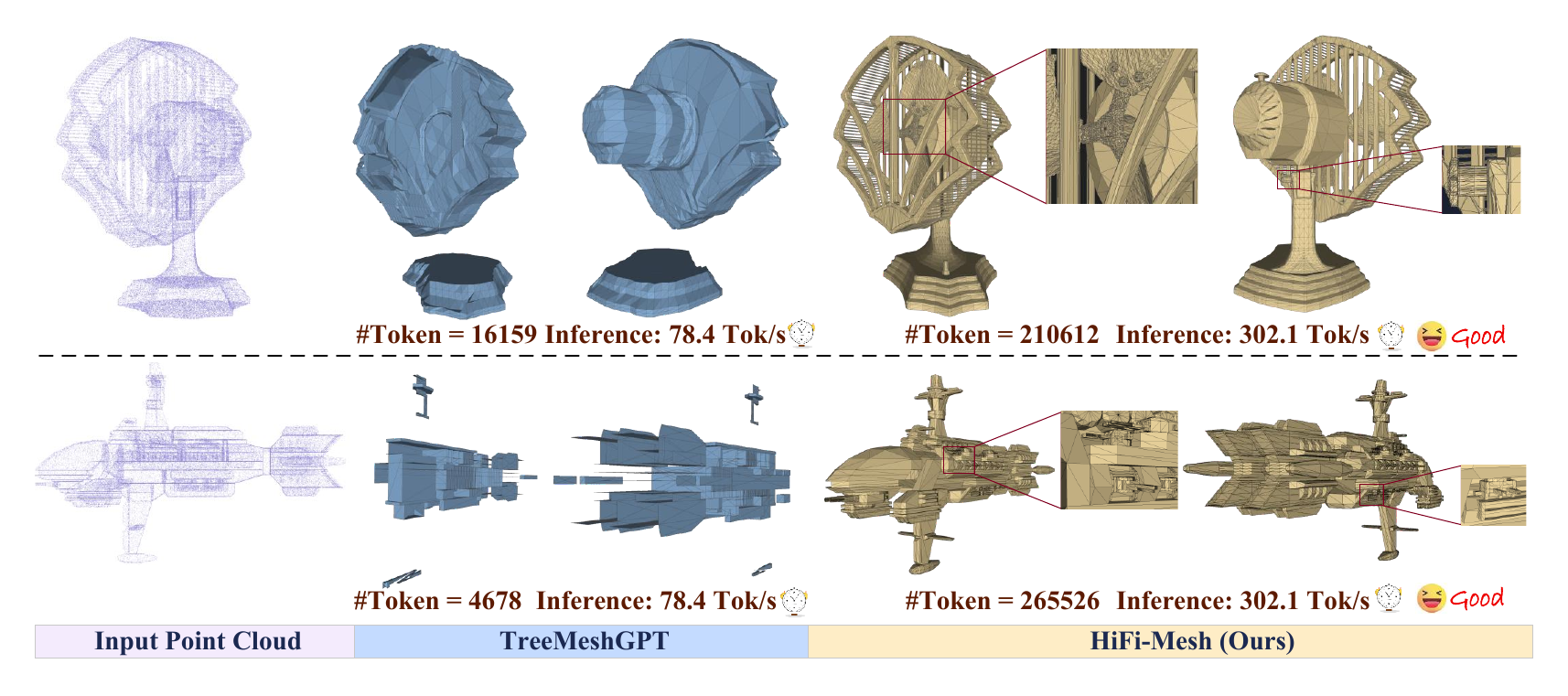}
	\end{overpic}
	\caption{Visual Comparisons between TreeMeshGPT \cite{lionar2025treemeshgpt} and HiFi-Mesh (Ours). A mesh can be tokenized into a 1D topological sequence. \#Token represents sequence length with the density of vertices and faces proportional to it. $X$ Tok/s denotes the inference speed of sequence token. HiFi-Mesh enables detail-rich mesh generation with faster speed.}
	\label{fig:homepage}
\end{figure*}

To address this challenge, recent research focuses on tokenizing 3D meshes into 1D sequences, using autoregressive models to generate them for direct modeling of mesh vertices and faces, thereby preserving topological information \cite{weng2024pivotmesh}. MeshXL \cite{chen2024meshxl} directly flattens vertices but results in an inefficient representation, reaching sequence lengths of $3vf$ ($f$: faces, $v$: vertices per face). MeshAnythingV2 \cite{chen2024meshanythingv2} introduced Adjacent Mesh Tokenization, compressing sequence representation and extending to arbitrary object generation. Although subsequent algorithms have continued to improve upon these foundations, two critical bottlenecks persist.

First, sequence compression algorithms have compression rate limits, and representation of high-fidelity meshes requires sequences on the order of hundreds of thousands. However, constrained by the computational overhead of autoregressive models, current methods are still limited to a maximum generatable length of $50,000$ tokens \cite{tang2024edgerunner}. This bottleneck restricts the structural detail of generated meshes.
Second, autoregressive model inference must follow strict forward recursive generation from the sequence beginning \cite{siddiqui2024meshgpt}. This serialized mechanism significantly constrains inference efficiency. To alleviate this bottleneck, existing methods truncate the context directly, retaining only a limited-length historical sequence as the inference prefix. However, such operations result in substantial information loss, causing significant performance degradation in later stages of long sequence inference.

To address these challenges, we propose HiFi-Mesh for efficient generation of detail-rich 3D meshes, which comprises a Latent Autoregressive Network (LANE) model and Adaptive Computation Graph Reconfiguration (AdaGraph) strategy.
The core of the LANE lies in partitioning topological sequences into multiple structural units and introducing sequence semantic reconstruction to build corresponding latent space representations from point cloud as autoregressive dependencies, while ensuring coherence and compactness of structural information. First, the model performs autoregressive generation following subsequence order, with each step depending only on current and preceding latent spaces, reducing long-sequence dependencies. Second, to enhance local modeling capability, we introduce learnable query tokens as structural queriers that dynamically perceive key features in latent spaces, achieving sequential alignment and detail restoration. This design reduces computational overhead while effectively improving maximum generatable length and expressible structural details.

To further enhance inference efficiency, we propose AdaGraph strategy. This strategy leverages the autoregressive dependencies introduced by LANE to achieve spatiotemporal decoupling in generation process. Specifically, AdaGraph first exploits the compactness of latent space representations for rapid hierarchical construction. Subsequently, it decomposes serial inference into multiple independent pathways, where each pathway dynamically reconstructs dedicated computational subgraphs based on target subsequences and selectively activates corresponding latent spaces for parallel generation. Compared to previous methods, AdaGraph maintains complete contextual information, avoiding performance degradation while achieving approximately $300\%$ speed improvement. Figures~\ref{fig:introduction_display} and~\ref{fig:homepage} shows performance and visual comparisons, respectively.

Our contributions are as follows:
\begin{itemize}
    \item We propose HiFi-Mesh, which incorporates compact autoregressive dependencies in sequence generation to enable rapid generation of detail-rich 3D meshes.
    \item LANE and AdaGraph are introduced to enhance maximum generatable sequence length and enable parallel inference, respectively.
    \item Compared to existing methods, HiFi-Mesh demonstrates superior performance in generation speed, structural detail, and geometric consistency with point clouds, providing valuable insights for addressing such challenges.
\end{itemize}
\begin{figure*}[!t]
	\centering
	\begin{overpic}[width=1.0\textwidth, trim=0 20 0 0, clip]{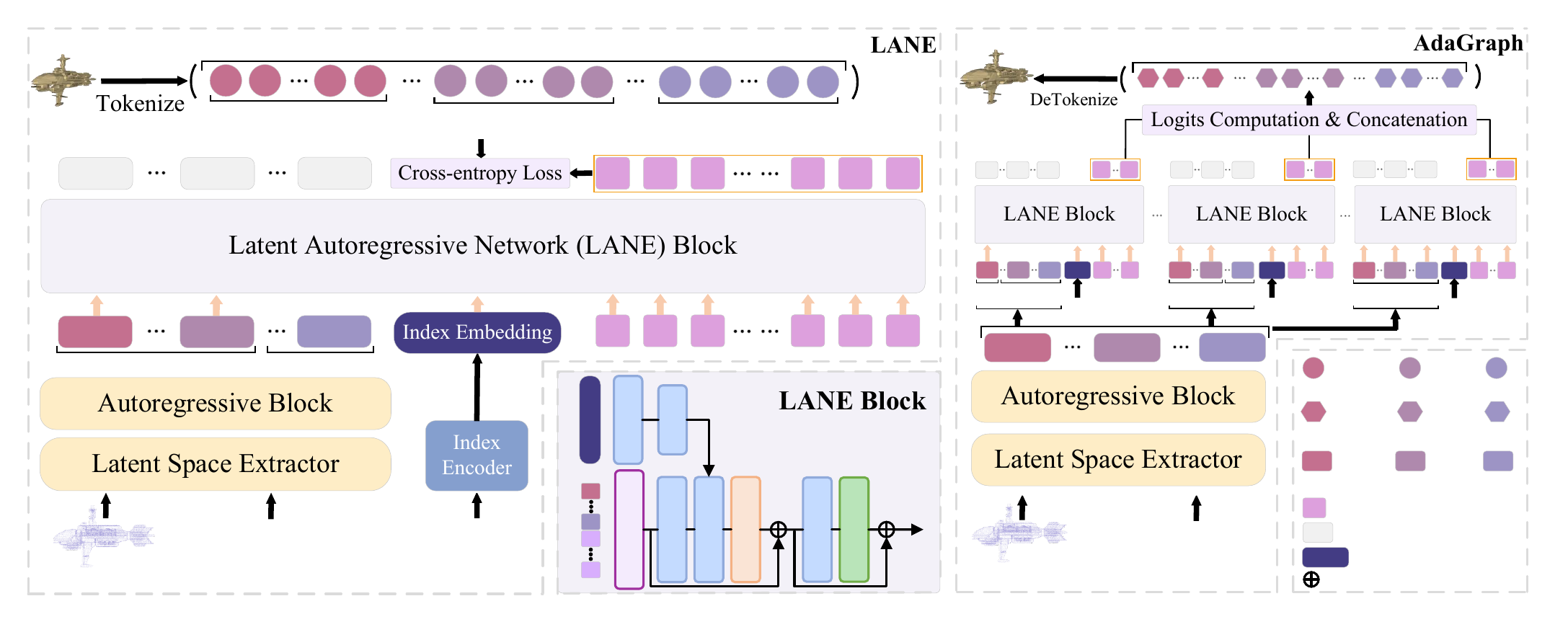}
		\put(63.45,17.8){\footnotesize\parbox{2cm}{\centering{{$I$:$1$}}}}
		\put(55.95,19.0){\scriptsize\parbox{2.5cm}{\centering{\textcolor{blockgray}{$1$}}}}
		\put(58.9,19.0){\scriptsize\parbox{2.5cm}{\centering{\textcolor{blockgray}{$M$-$1$}}}}
		\put(70.8,19.0){\scriptsize\parbox{2cm}{\centering{\textcolor{blockgray}{$m$}}}}
		\put(73.5,19.0){\scriptsize\parbox{2cm}{\centering{\textcolor{blockgray}{$M$-$m$}}}}
		\put(42.6,15.6){\scriptsize\parbox{2cm}{\centering{\textcolor{black}{$Q$}}}}
		\put(76.7,17.8){\footnotesize\parbox{2cm}{\centering{{$I$:$m$}}}}
		\put(83.7,19.0){\scriptsize\parbox{2cm}{\centering{\textcolor{blockgray}{$M$}}}}
		\put(88.2,17.8){\footnotesize\parbox{2cm}{\centering{{$I$:$M$}}}}
		\put(13.3,2.2){\scriptsize\parbox{1.5cm}{\centering{{Sequence Lehgth ($L$): $l$}}}}
		\put(26,2.2){\scriptsize\parbox{1.6cm}{\centering{{Subsequence Index ($I$): $m$}}}}
		\put(72.5,2.2){\scriptsize\parbox{1.5cm}{\centering{{Sequence Lehgth ($L$): $l$}}}}
		\put(-2.8,23.7){\large\parbox{2.5cm}{\centering{\textcolor{blockK}{K\textbf{×}}}}}
		\put(3,14.3){\footnotesize\parbox{2.5cm}{\centering{\textcolor{blockgray}{$m$}}}}
		\put(13.4,14.3){\footnotesize\parbox{2.5cm}{\centering{\textcolor{blockgray}{$M$-$m$}}}}
		\put(-0.8,16.5){\small\parbox{2.5cm}{\centering{{$sc_1$}}}}
		\put(7,16.5){\small\parbox{2.5cm}{\centering{{$sc_m$}}}}
		\put(14.3,16.5){\small\parbox{2.5cm}{\centering{{$sc_M$}}}}
		\put(17.5,34.8){\scriptsize\parbox{5.5cm}{\centering{{Sequence $S$ ($M$ Subsequence, Length: $l$)}}}}
		\put(24.5,29.9){\scriptsize\parbox{3cm}{\centering{{$m^{th}$ Subsequence ($s_m$)}}}}
		\put(41.5,28.6){\small\parbox{2.5cm}{\centering{{$\hat{s_m}$}}}}
		\put(64.0,28.3){\scriptsize\parbox{2.5cm}{\centering{{$\hat{s_1}$}}}}
		\put(75.5,28.3){\scriptsize\parbox{2.5cm}{\centering{{$\hat{s_m}$}}}}
		\put(86.9,28.3){\scriptsize\parbox{2.5cm}{\centering{{$\hat{s_M}$}}}}
		\put(65.5,34.8){\scriptsize\parbox{5.5cm}{\centering{{ Predictd Sequence $\hat{S}$ (Length: $l$)}}}}
		\put(56.1,25.2){\scriptsize\parbox{2.5cm}{\centering{\textcolor{blockK}{K\textbf{×}}}}}
		\put(65.1,25.2){\scriptsize\parbox{2.5cm}{\centering{$lk_1$}}}
		\put(76.9,25.2){\scriptsize\parbox{2.5cm}{\centering{$lk_m$}}}
		\put(88.5,25.2){\scriptsize\parbox{2.5cm}{\centering{$lk_M$}}}
		\put(68.4,25.2){\scriptsize\parbox{2.5cm}{\centering{\textcolor{blockK}{K\textbf{×}}}}}
		\put(80.3,25.2){\scriptsize\parbox{2.5cm}{\centering{\textcolor{blockK}{K\textbf{×}}}}}
		\put(45.5,17.3){\footnotesize\rotatebox{270}{\parbox{2.5cm}{\centering{{$\alpha$,$\beta$}}}}}
		\put(0.2,0.9){\scriptsize\parbox{2.5cm}{\centering{{Point Cloud ($P$)}}}}
		\put(59.1,0.9){\scriptsize\parbox{2.5cm}{\centering{{Point Cloud ($P$)}}}}
		\put(58.0,15.5){\small\parbox{2.5cm}{\centering{{$sc_1$}}}}
		\put(65.1,15.5){\small\parbox{2.5cm}{\centering{{$sc_m$}}}}
		\put(71.8,15.5){\small\parbox{2.5cm}{\centering{{$sc_M$}}}}
		\put(81.3,12.8){\scriptsize\parbox{2.5cm}{\centering{{Subsequence Token}}}}
		\put(80.4,9.8){\scriptsize\parbox{2.5cm}{\centering{{Predicted Token}}}}
		\put(81.7,6.7){\scriptsize\parbox{3cm}{\centering{{Hierarchical Latent Space}}}}
		\put(84.7,5.2){\scriptsize\parbox{2.5cm}{\centering{{Learnable Query}}}}
		\put(84.0,3.9){\scriptsize\parbox{2.5cm}{\centering{{Unused Token}}}}
		\put(85.0,2.15){\scriptsize\parbox{2.5cm}{\centering{{Index Embedding}}}}
		\put(83.3,0.9){\scriptsize\parbox{2.5cm}{\centering{{Summation}}}}
		\put(39.6,10.9){\rotatebox[origin=c]{270}{\scriptsize\parbox{2.5cm}{\centering{{Condition}}}}}
		\put(42.5,10.9){\rotatebox[origin=c]{270}{\scriptsize\parbox{2.5cm}{\centering{{MLP}}}}}
		\put(39.4,3.9){\rotatebox[origin=c]{270}{\scriptsize\parbox{2.5cm}{\centering{{Input Tokens}}}}}
		\put(42.2,3.9){\rotatebox[origin=c]{270}{\scriptsize\parbox{2.5cm}{\centering{{LayerNorm}}}}}
		\put(44.7,3.9){\rotatebox[origin=c]{270}{\scriptsize\parbox{2.5cm}{\centering{{Scale, Shift}}}}}
		\put(47.2,3.9){\rotatebox[origin=c]{270}{\scriptsize\parbox{2.5cm}{\centering{{LANE-Attn}}}}}
		\put(51.55,3.9){\rotatebox[origin=c]{270}{\scriptsize\parbox{2.5cm}{\centering{{LayerNorm}}}}}
		\put(54.05,3.9){\rotatebox[origin=c]{270}{\scriptsize\parbox{2.5cm}{\centering{{FFN}}}}}
	\end{overpic}
	\caption{The illustration of our proposed method. During training, the purpose of LANE model (\textbf{\textit{left}}) uniformly divides the topological sequence into $M$ subsequences for sequential generation. The model takes point cloud $P$ and sequence length $L$ as inputs, employing a Latent Space Extractor and Autoregressive Block for sequence semantic reconstruction to build hierarchical latent space representations $\{sc_m\}_{m\in[1,M]}$ as compact autoregressive dependencies. Subsequently, $K$ LANE Blocks serve as generation components, utilizing Learnable Queries $Q$ as latent space queriers. Guided by subsequence index $I=\{m\} _{m\in[1,M]})$, each step generates the $m^{th}$ subsequence $\hat{s_m}$ from the only first $m$ latent spaces. During inference, the AdaGraph strategy (\textbf{\textit{right}}) rapidly generates hierarchical latent space representations $\{sc_m\}_{m\in[1,M]}$ and instantiates $M$ independent pathways $\{lk_m\}_{m\in[1,M]}$ composed of LANE Blocks. Each pathway dynamically reconstructs a dedicated computational subgraph based on the target subsequence and selectively activates corresponding latent spaces, enabling fast parallel generation. Finally, mesh generation is completed through subsequence concatenation and detokenization operations.}
	\label{fig:framework}
\end{figure*}
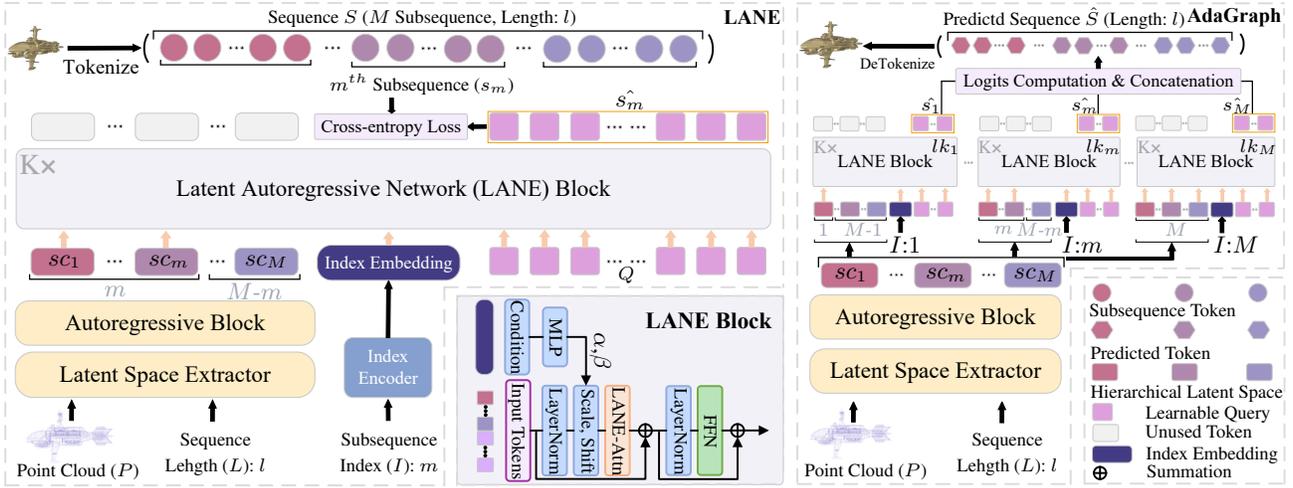
\section{2\ \ \ \  Related Work}
\subsection{2.1\ \ \ \ 3D Generation}
Traditional rule-based modeling techniques \cite{wu2024rg} have gradually been replaced by diverse learned representations, including voxel generation \cite{zhao2025hunyuan3d, chen20253dtopia}, implicit function encoding \cite{lim2024tifu, di2025hyper}, and direct mesh construction methods \cite{zhou2024simple, wu2024direct3d}. Regarding model architectures, Generative Adversarial Networks \cite{wang2024msg} are giving way to diffusion models \cite{li2025step1x, li2025cmd} and specialized end-to-end frameworks \cite{tang2024lgm}, which exhibit superior capabilities in generating 3D assets. These advancements provide a more efficient methodological foundation for 3D content creation.
\subsection{2.2\ \ \ \ 3D Mesh Generation}
Meshes precisely describe object surface geometry \cite{liu2024meshformer}. Early research used Gaussian splatting to predict 3D meshes from 2D images \cite{hu2025turbo3d, huang2025spar3d}, but 2D images lack depth and geometric details, creating an information bottleneck. 
To address this, researchers generate SDF mesh representations directly \cite{zhang20233dshape2vecset, hunyuan3d2025hunyuan3d, xiang2025structured}. While these methods reduce information loss, their reliance on fixed thresholds causes geometric discontinuities.
Recent studies tokenize 3D meshes into 1D sequences, using autoregressive models to directly model mesh vertices and faces. Methods like VAT and MeshAnything use VQVAE for tokenization \cite{zhang20243d, chen2024meshanything}, but suffer from over-compression and significant geometric detail loss. Alternatively, MeshArt and Mesh-RFT directly tokenize meshes into topological sequences with specific ordering \cite{gao2025meshart, liu2025mesh}, preserving complete geometric and topological information. However, while achieving high information fidelity, the massive sequence scale for detail-rich meshes creates prohibitive computational overhead for autoregressive models.
\subsection{2.3\ \ \ \ Autoregressive Mesh Generation}
When processing detail-rich meshes, significant computational efficiency challenges arise. While BPT and EdgeRunner reduced sequence lengths through improved tokenization \cite{weng2025scaling, tang2024edgerunner}, these meshes still require hundreds of thousands of tokens, exceeding current computational limits. 
Meshtron \cite{hao2024meshtron} addressed this with sequence fragment-based adjacency structure prediction. Although this significantly increased the scale of generatable meshes, it further reduced inference speed, which was already constrained by forward recursive generation mechanisms, showing approximately 30\% slower compared to TreeMeshGPT \cite{lionar2025treemeshgpt}. In this work, we propose a novel framework that reduces computational overhead to handle longer sequences. Crucially, we further overcome inference bottlenecks, enabling rapid generation of detail-rich 3D meshes.
\section{3\ \ \ \ Method}
\subsection{3.1\ \ \ \ Latent Autoregressive Network}
The Latent Autoregressive Network (LANE) model (Figure~\ref{fig:framework}, \textbf{\textit{left}}) uniformly splits the topological sequence $S$ into $M$ consecutive subsequences $\{s_m\}_{m \in [1,M]}$ and performs sequential generation conditioned on subsequence indexes $I=\{m\}_{m \in [1,M]}$, sequence length $L$, and point cloud $P$. The core principle of model is to accurately reconstruct local geometric structures into corresponding latent spaces while ensuring structural coherence. The model then uses compact latent space representations to replace the long historical token sequences required for prediction, thereby reducing computational cost, improving structural detail expressiveness. Specifically, the model consists of three components: Latent Space Extractor, Autoregressive Block, LANE block.
\paragraph{Latent Space Extractor.}
The Latent Space Extractor employs a Point Cloud Encoder to extract highly compact structural features from high-resolution input point clouds $P$, then reconstructs fine-grained features through learnable Upsample modules that gradually decode the compressed information. Compared to previous methods, our input point clouds have $\mathbf{16}$ times higher resolution, containing richer geometric details that establish a foundation for enhanced structural detail expressiveness.
Specifically, we perform four random samplings, $\{X_i\}_{i \in [1,4]}$, from the input mesh surface, containing $\{N_i\}_{i \in [1,4]}$ points respectively, where $X_i \in R^{N_i\times3}$, and $N_2<N_3<N_4<N_1$. $X_1$ serves as the input point cloud $P$ and is encoded into a latent code by the Point Cloud Encoder built from cross-attention, using $X_2$ as query. Subsequently, $X_3$ and $X_4$ act as queries in cross-attention-based Upsample modules to progressively obtain a more refined latent code $Z$.

Additionally, this Extractor includes a Latent Space Constructor that first initializes corresponding latent spaces in subsequence order, then leverages the sequence length $L$ to guide the latent spaces in accurately partitioning subsequence structure boundaries in the latent code, achieving precise extraction and encoding of target features. The construction process of $m^{th}$ latent space is defined as:
\begin{equation}
	\begin{aligned}
		 sc_m^e = CrossAttn([sc_m^{init};L_e],Z),
	\end{aligned}
	\label{eq:latent_space_constructor}
\end{equation}
where, $CrossAttn(\cdot,\cdot)$ denotes the cross-attention layer with the left side as Query. ${sc_m^{init}}$ represents the $m^{th}$ initialized latent space, where $m\in[1,M]$. $L_e$ and $Z$ denote the embedding of sequence length $L$ and the latent code reconstructed by the Upsample modules, respectively. $[\ ;\ ;\ ]$ indicates vertical concatenation of multiple information sources. Detailed designs are presented in technical appendix.
\paragraph{Autoregressive Block.}
The core of this component is to transform the encoded latent spaces $\{sc_{m}^e\}_{m\in[1,M]}$ into hierarchical latent space representations $\{sc_m\}_{m\in[1,M]}$ to capture dependencies between subsequences and maintain global consistency of topological structure. Specifically, this component utilizes an autoregressive mechanism to sequentially process $\{sc_{m}^e\}_{m\in[1,M]}$ generated by the Latent Space Extractor, thereby constructing $\{sc_m\}_{m\in[1,M]}$ that reflects the hierarchical nature of the overall topological structure. This representation preserves local detail information while ensuring global coherence, providing a high-quality feature representation foundation for subsequent sequence generation tasks. Its expression is as follows:
\begin{equation}
	\begin{aligned}
		[sc_1; sc_2; ...; sc_M] = CausalAttn([sc_1^e; sc_2^e; ...; sc_M^e]),
	\end{aligned}
	\label{eq:Autoregressive Block}
\end{equation}
where, $CausalAttn(\cdot)$ represents the causal attention mechanism in autoregressive models, $[\ ;...\ ;]$ denotes vertical concatenation of multiple information sources.
\paragraph{LANE Block.}
The LANE Block, denoted as $\mathcal{E}_{block}$, aims to sequentially generate subsequences from the constructed hierarchical latent space representations $\{sc_m\}_{m\in[1,M]}$. Specifically, learnable query tokens $Q$ are introduced as queriers. When generating the $m^{th}$ subsequence, coherent structural features are extracted from the latent spaces $\{sc_1, sc_2,…,sc_{\mathbf{m-1}}\}$ corresponding to preceding subsequences, thereby eliminating dependence on long historical token sequences. Subsequently, with the aid of subsequence index $I$, $Q$ further perceives structural details from the current corresponding latent space $\{sc_{\mathbf{m}}\}$, enhancing local modeling capability. The interaction can be defined as:
\begin{equation}
	\begin{aligned}
		V_{out} = \mathcal{E}_{block}(\{sc_1, sc_2,…,sc_{\mathbf{m}}\}, I_m^e, Q)
	\end{aligned}
	\label{eq:lane_block}
\end{equation}
where, $I_m^e$ represents the embedding of subsequence index when $I$ takes the value $m$. The forward pass of $\mathcal{E}_{block}$ is illustrated in Figure~\ref{fig:framework}, where the specific design and definition of LANE-Attn to utilize $\{sc_m\}_{m\in[1,M]}$ will be presented in technical appendix. This block is repeated $K$ times.
\paragraph{Loss Function.}
LANE model, denoted as $\mathcal{E}$, employs cross-entropy loss to predict $m^{th}$ subsequence tokens:
\begin{equation}
	\begin{aligned}
		&\mathcal{L}_{ce} = CrossEntropy(\hat{s_m}, s_m), \\
		&\hat{s_m} = \mathcal{E}(P, L, I=m),
	\end{aligned}
	\label{eq:loss_function}
\end{equation}
where $\hat{s_m}$ and $s_m$ denotes the predicted classification logits subsequence and the ground truth subsequence, respectively.
\subsection{3.2\ \ Adaptive Computation Graph Reconfiguration}
The AdaGraph strategy (Figure~\ref{fig:framework}, \textbf{\textit{right}}), leverages the compact autoregressive dependencies introduced by LANE model to achieve spatiotemporal decoupling in generation process, thereby further improving inference speed. Specifically, this strategy comprises two core stages: Hierarchical Latent Space Construction and Subgraph Reconstruction.
\begin{table*}[!t]
	\scriptsize
	\centering
	\renewcommand{\arraystretch}{1.0}
	\resizebox{\textwidth}{!}{
		\begin{tabular}{lccccclcc}
			\toprule
			\multirow{4}{*}{\textbf{Methods/Metrics}} & \multicolumn{5}{c}{\textbf{Generation Metric}}                                                            &  & \multicolumn{2}{c}{\textbf{Efficiency Metric}}              \\ \cline{2-6} \cline{8-9} 
			& \textbf{Chamfer $\downarrow$}              & \textbf{Normal $\downarrow$}               & \multicolumn{2}{c}{\textbf{Point-to-Mesh}} & \textbf{MOS $\uparrow$}                 &  & \textbf{Length $\uparrow$}                    & \textbf{Inference $\uparrow$}              \\ \cline{4-5}
			& \multirow{2}{*}{($\times 10^{-1}$)} & \multirow{2}{*}{($\times 10^{-1}$)} & \textbf{Mean $\downarrow$}          & \textbf{Hausorff $\downarrow$}         & \multirow{2}{*}{\%} &  & \multirow{2}{*}{\#Token} & \multirow{2}{*}{Tok/s} \\ \cline{4-5}
			&                      &                      & \multicolumn{2}{c}{($\times 10^{-2}$)}      &                     &  &                           &                        \\ \cline{1-6} \cline{8-9} 
			TreeMeshGPT \cite{lionar2025treemeshgpt}                     & 3.413                & \textbf{2.001}                & 3.242         & 8.253        & 5.26                &  & 30,000                     & \underline{78.4}                   \\
			MeshAnythingV2 \cite{chen2024meshanything}                   & \underline{0.501}                & 4.951                & 2.631         & 6.374        & 3.72                &  & 10,083                     & 67.3                   \\
			EdgeRunner \cite{tang2024edgerunner}                      & 0.532                & 6.339                & \underline{0.921}         & \underline{3.764}        & \underline{9.31}                &  & \underline{49,406}                     & 52.8                   \\
			HiFi-Mesh (Ours)                      & \textbf{0.075}                & \underline{3.621}                & \textbf{0.638}         & \textbf{2.251}        & \textbf{81.17}               &  & \textbf{300,000}                    & \textbf{302.1}                  \\
			\bottomrule
		\end{tabular}
	}
	\caption{Quantitative comparisons on seven metrics. \textbf{Bold} and \underline{underlined} indicate the best and second-best performance, respectively. Despite minor Normal Consistency metric limitations, HiFi-Mesh excels in mesh consistency and inference speed. User studies confirm the aesthetic and detailed quality of generated meshes. Around $300\%$ boost in inference speed enables processing $6\times$ longer sequences, achieving high-fidelity 3D mesh generation. All test results are obtained using a single H20.}
	\label{tab:quantitative comparisons}
\end{table*}
\subsubsection{Hierarchical Latent Space Construction.} Using the Latent Space Extractor and Autoregressive Block, the hierarchical latent space representations are rapidly constructed from the input point cloud $P$ under the guidance of sequence length $L$. This process is denoted as $\mathcal{T}$ and defined as:
\begin{equation}
	\begin{aligned}
		[sc_1; sc_2; ...; sc_M]=\mathcal{T}(P, L).
	\end{aligned}
	\label{eq:LatPar_T}
\end{equation}
This process follows the definitions in Equations \ref{eq:latent_space_constructor} and \ref{eq:Autoregressive Block}.
\subsubsection{Subgraph Reconstruction.}
LANE internally instantiates $M$ independent pathways $\{lk_m\}_{m\in[1,M]}$, with each pathway responsible for generating specific subsequences. Each pathway dynamically reconstructs its dedicated computational subgraph and selectively activates corresponding latent spaces, achieving spatiotemporal decoupling in the generation process. This strategy employs batch processing to coordinate multi-pathway collaboration, leveraging the advantages of modern parallel computing architectures to enable efficient parallel generation, significantly improving overall inference speed. This process is represented as:
\begin{equation}
	\begin{aligned}
		&\hat{S} = [\hat{s_1}; ...; \hat{s_m}; ...; \hat{s_M}], \\
		&\hat{s_m} = lk_m(\{sc_1, sc_2, ..., sc_{m}\}, I=m, Q),
	\end{aligned}
	\label{eq:LatPar}
\end{equation}
where each pathway consists of $K$ LANE Blocks, and the generation process follows Equation \ref{eq:lane_block}. Finally, by concatenating all predicted classification logic subsequences $\{\hat{s_m}\}_{m\in[1,M]}$ and performing detokenization operation on $\hat{S}$, the 3D mesh is generated.
\begin{figure*}[!t]
	\centering
	\begin{overpic}[width=1.0\textwidth, trim=0 20 0 0, clip]{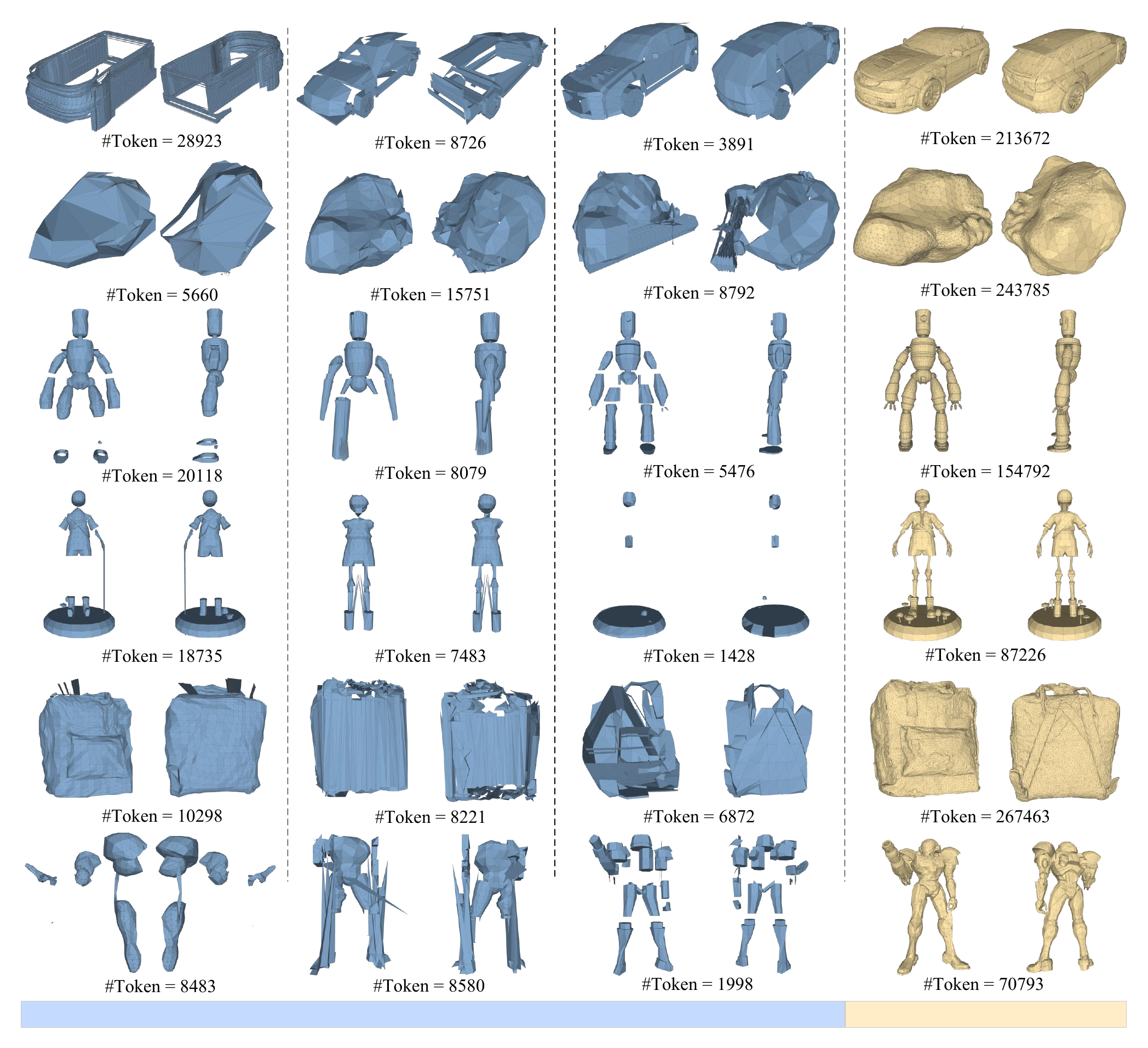}
		\put(-0.5,0.6){\parbox{5cm}{\footnotesize\centering{{\textbf{TreeMeshGPT}: 78.4 Token/s}}}}
		\put(25.5,0.6){\footnotesize\parbox{4cm}{\centering{{\textbf{EdgeRunner}: 52.8 Token/s}}}}
		\put(47.5,0.6){\footnotesize\parbox{5cm}{\centering{{\textbf{MeshAnythingV2}: 67.3 Token/s}}}}
		\put(76.0,0.6){\footnotesize\parbox{4cm}{\centering{{\textbf{HiFi-Mesh}: 302.1 Token/s}}}}
	\end{overpic}
	\caption{Qualitative comparisons between TreeMeshGPT \cite{lionar2025treemeshgpt}, EdgeRunner \cite{tang2024edgerunner}, MeshAnythingV2 \cite{chen2024meshanything} and HiFi-Mesh (Ours). MeshAnythingV2 exhibits limitations in maximum sequence scale, causing most significant structural information loss. EdgeRunner shows theoretical improvements but maintains preference for small-scale sequences. TreeMeshGPT achieves some improvement in detail modeling but still suffers from insufficient sequence generation capability. Finally, HiFi-Mesh enables superior preservation of details and generation of detail-rich meshes.}
	\label{fig:qualitative_comparisons}
\end{figure*}
\section{4\ \ \ \ Experiment}
\subsection{4.1\ \ \ \ Implementation}
\subsubsection{Training Details.} The experiments are implemented using PyTorch, conducted on $8$ NVIDIA H20 GPUs. We employ the Adam optimizer and optimize for six weeks, with a learning rate decay from $1e$-$4$ to $1e$-$6$. Flash Attention \cite{shah2024flashattention} is integrated to optimize computation. All the results are tested on a single H20 GPU. Model training utilize the Objaverse dataset \cite{objaverseXL,tang2024lgm}.
\subsubsection{Tokenization Algorithm.} This paper leverages tokenization algorithm of EdgeRunner \cite{tang2024edgerunner} for autoregressive model compatibility, optimizing the balance between geometric fidelity and compression ratio. The topological sequence labeling is constructed by quantizing coordinate outputs into $512$ discrete categories. 
\begin{figure*}[h]
	\centering
	\begin{overpic}[width=1.0\textwidth, trim=0 17 0 0, clip]{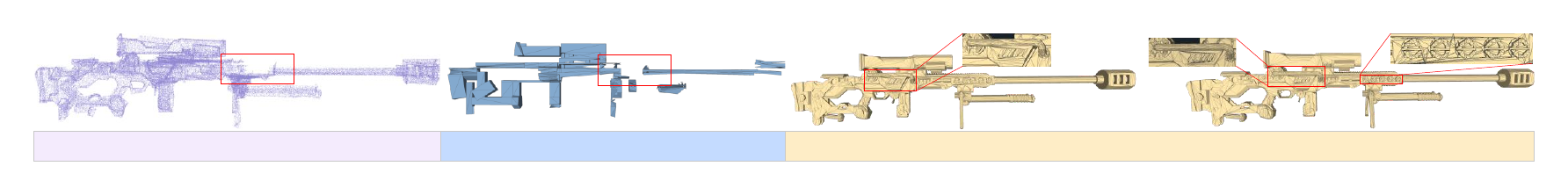}
		\put(2,0.5){\footnotesize\parbox{4cm}{\centering{{\textbf{Input Point Cloud}}}}}
		\put(25.0,0.5){\footnotesize\parbox{5cm}{\centering{{\textbf{EdgeRunner} (\#Token = 4.9K)}}}}
		\put(60.5,0.5){\footnotesize\parbox{5cm}{\centering{{\textbf{HiFi-Mesh} (\#Token = 67K/109K)}}}}
	\end{overpic}
	\caption{Functional presentation. Point clouds are sampled from imperfect meshes. HiFi-Mesh generates meshes with rich details guided by different sequence lengths, while EdgeRunner \cite{tang2024edgerunner} faces local topological breaks.}
	\label{fig:functional_presentation}
\end{figure*}
\begin{figure}[!t]
	\centering
	\begin{overpic}[width=0.49\textwidth, trim=17 15 0 20, clip]{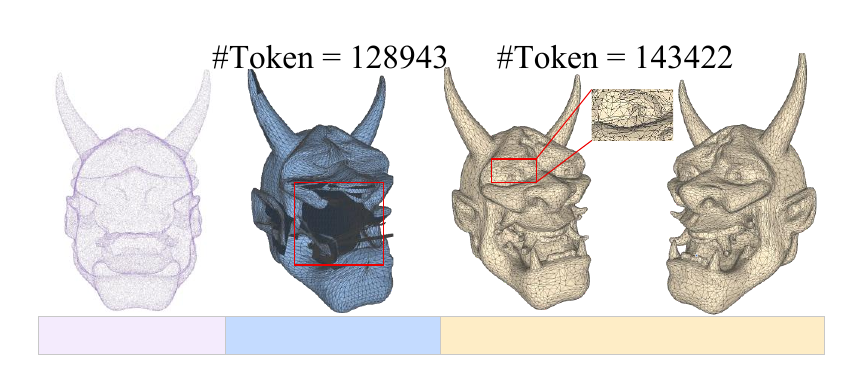}
		\put(-10.5,2.0){\footnotesize\parbox{4cm}{\centering{{\textbf{Input}}}}}
		\put(13.3,2.0){\footnotesize\parbox{4cm}{\centering{\textbf{w/o $\boldsymbol{\{sc_m\}_{[1,M]}}$ }}}}
		\put(52,2.0){\footnotesize\parbox{4cm}{\centering{\textbf{HiFi-Mesh (Ours)}}}}
	\end{overpic}
	\caption{Ablation study. (w/o) $\{sc_m\}_{[1,M]}$ represents without constructing hierarchical latent space representations in HiFi-Mesh. It leads to local topological breaks, geometric integrity loss, and inability to express detailed structures.}
	\label{fig:LANE_effectiveness}
\end{figure}
\subsection{4.2\ \ \ \ Comparisons}
\label{sec:quantitative comparisons}
Three algorithms are selected as comparions: MeshAnythingV2 \cite{chen2024meshanything}, EdgeRunner \cite{tang2024edgerunner}, and TreeMeshGPT \cite{lionar2025treemeshgpt}. Four generative metrics are employed to evaluate mesh generation quality: Chamfer Distance (CD), Normal Consistency (NC), Point-to-Mesh Distance (PMD), and MOS. Two efficiency metrics are used to assess model efficiency: Maximum Generateble Sequence Length (ML), and Inference Speed (IS). CD evaluates spatial proximity between reconstructed and original mesh. NC assesses the consistency of normal vectors between adjacent faces. PMD measures the distance from points in the point cloud to the reconstructed mesh surface, evaluated from both Mean distance and Hausdorff distance perspectives. MOS reflects the proportion of participants favoring a specific model. To collect objective evaluations, ten users are recruited for the assessment task. Length indicates the maximum sequence length that the model can generate. Inference represents the number of token that can be generated per second. The generative metrics are implemented by the PyTorch3D \cite{ravi2020accelerating}.
\subsubsection{Quantitative Comparisons.}
The complete comparison results are shown in Table~\ref{tab:quantitative comparisons}. In terms of generation quality, except for a slight deficiency in Normal Consistency, HiFi-Mesh demonstrates superior performance in mesh consistency and robustness (Chamfer Distance and Point-to-Mesh Distance). User preference studies further validate the aesthetic value and expressible structural detail of our generated meshes. Regarding efficiency of resource utilization, we enable a $6\times$ increase in maximum generatable sequence length and further leverage the AdaGraph strategy to achieve a $300\%$ improvement approximately in inference speed, allowing rapid generation of detail-rich 3D meshes.
\subsubsection{Qualitative Comparisons.}
Figure~\ref{fig:qualitative_comparisons} shows visual comparisons between HiFi-Mesh and three baseline methods \cite{lionar2025treemeshgpt, tang2024edgerunner, chen2024meshanything}. Sequence generation scale directly affects mesh reconstruction capability. MeshAnythingV2 suffers from severe limitations in maximum generatable sequence scale, leading to most  significant structural information loss. While EdgeRunner shows theoretical improvements in generation length, it still exhibits a strong preference for small-scale sequences in practical applications. TreeMeshGPT achieves relatively balanced performance and can generate larger sequences, resulting in some improvement in mesh details.
However, the maximum sequence generation capabilities of these three methods remain insufficient to fully capture the geometric features of complex surfaces, limiting mesh structural detail expressiveness and generation stability. In contrast, HiFi-Mesh significantly increases the upper limit of sequence generation scale, enabling the model to more effectively learn and preserve surface detail information, thereby achieving stable generation of detail-rich 3D meshes.
\subsubsection{Functional Presentation.} As shown in Figure~\ref{fig:functional_presentation}, introducing surface damage to input meshes during training not only enhances model robustness but also enables HiFi-Mesh to perform mesh repair tasks: sampling point clouds from defective meshes and then reconstructing detailed, high-quality meshes. Additionally, by incorporating the sequence length $L$ as a control parameter, HiFi-Mesh achieves length control over the generation process.
\subsection{4.3\ \ \ \ Ablation Study}
\subsubsection{Components Effectiveness.}
As shown in Figure~\ref{fig:LANE_effectiveness}. (w/o) $\{sc_m\}_{[1,M]}$ denotes the directly use of global geometric features extracted by the Point Cloud Encoder to generate target subsequences without constructing hierarchical latent space representations in HiFi-Mesh. Experimental results show that the absence of coherent prior structural features and local detail information provided by latent spaces leads to local topological breaks, geometric integrity loss, and inability to express detailed structures in the generated results.

Table~\ref{tab:effectiveness} quantifies the importance of hierarchical latent space representations and AdaGraph strategy. Hierarchical latent spaces effectively capture local detail features while maintaining global geometric coherence, significantly improving the generation performance of the model. Additionally, the AdaGraph strategy leverages modern parallel computing architectures to further achieve approximately $3\times$ inference acceleration. Together, they validate the dual advantages of the proposed method in both efficiency and quality. More experiments are presented in technical appendix.
\begin{table}[!t]
	\footnotesize
	\centering
	\renewcommand{\arraystretch}{1.1}
	\resizebox{\columnwidth}{!}{
	\begin{tabular}{cccccc}
		\toprule
		\multirow{2}{*}{$\boldsymbol{\{sc_m\}_{[1,M]}}$} & \textbf{Chamfer$\downarrow$} & \textbf{Normal$\downarrow$} &  & \multirow{2}{*}{\textbf{AdaGraph}} & \textbf{Inference$\uparrow$} \\
		& ($10^{-1}$)     & ($10^{-1}$)    &  &                           & Tok/s     \\ \cline{1-3} \cline{5-6} 
		\ding{55}                   & 1.291   & 6.339  &  & 	\ding{55}                         & 121.7     \\
		\ding{51}                   & \textbf{0.157}   & \textbf{1.972}  &  & 	\ding{51}                         & \textbf{302.1}     \\ \bottomrule
	\end{tabular}
	}
	\caption{Ablation study. $\{sc_m\}_{[1,M]}$ represents the construction of hierarchical latent space representations in our method HiFi-Mesh. The hierarchical latent spaces provide compact autoregressive dependencies, significantly improving the generative performance of HiFi-Mesh. Additionally, the AdaGraph strategy is introduced to further achieve approximately $3\times$ inference acceleration.}
	\label{tab:effectiveness}
\end{table}

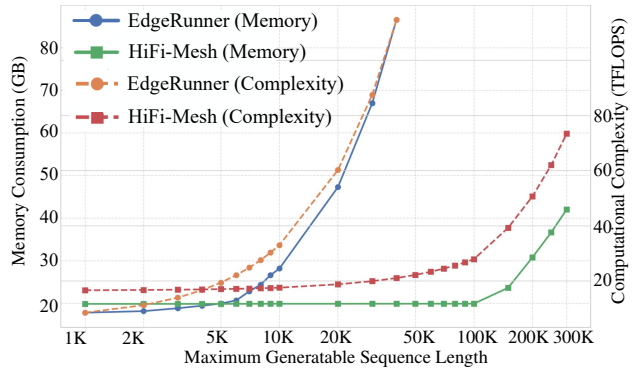
\begin{figure}[!t]
	\centering
	\begin{overpic}[width=0.48\textwidth, trim=0 10 0 15, clip]{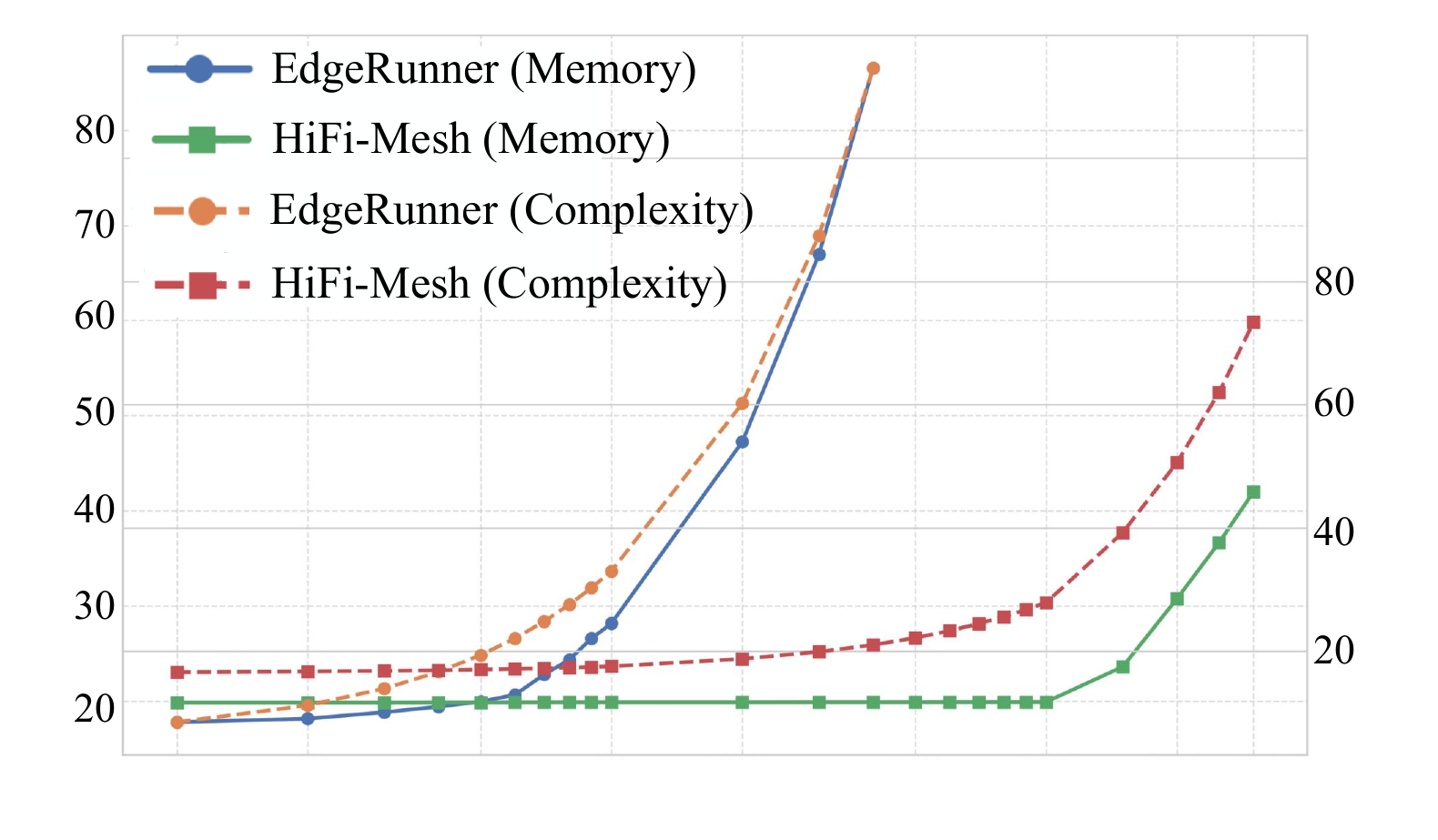}
	\put(59,2.2){\scriptsize\parbox{5cm}{\centering{{{300K}}}}}
	\put(52,2.2){\scriptsize\parbox{5cm}{\centering{{{200K}}}}}
	\put(44,2.2){\scriptsize\parbox{5cm}{\centering{{{100K}}}}}
	\put(35,2.2){\scriptsize\parbox{5cm}{\centering{{{50K}}}}}
	\put(22,2.2){\scriptsize\parbox{5cm}{\centering{{{20K}}}}}
	\put(13,2.2){\scriptsize\parbox{5cm}{\centering{{{10K}}}}}
	\put(3.5,2.2){\scriptsize\parbox{5cm}{\centering{{{5K}}}}}
	\put(8,2.2){\scriptsize\parbox{2cm}{\centering{{{2K}}}}}
	\put(-1,2.2){\scriptsize\parbox{2cm}{\centering{{{1K}}}}}
	\put(22,-1.2){\scriptsize\parbox{5cm}{\centering{{{Maximum Generatable Sequence Length}}}}}
	\put(0.8,29.7){\rotatebox[origin=c]{90}{\scriptsize\parbox{5cm}{\centering{{Memory Consumption (GB)}}}}}
	\put(94.2,29.7){\rotatebox[origin=c]{90}{\scriptsize\parbox{5cm}{\centering{{Computational Complexity (TFLOPS)}}}}}
	\end{overpic}
	\caption{Memory consumption and computational complexity for varying maximum generatable sequence lengths. HiFi-Mesh demonstrates gains when length exceeds $5K$, highlighting superior efficiency for sequence processing.}
	\label{fig:resource_occupation}

\end{figure} 

\subsubsection{Resource Consumption.}
Figure~\ref{fig:resource_occupation} presents a comparative analysis of resource consumption between HiFi-Mesh and traditional prediction methods of leveraging long historical sequence during training. The experiment uses a batch size of $4$, with EdgeRunner \cite{tang2024edgerunner} as the traditional baseline, evaluating computational complexity and memory consumption of both approaches across different maximum generatable sequence lengths.
Results show that HiFi-Mesh does not demonstrate efficiency advantages over EdgeRunner for short sequences, primarily due to additional computational overhead from constructing hierarchical latent space representations. However, when maximum sequence length exceeds $5,000$, HiFi-Mesh begins showing significant efficiency advantages that become increasingly pronounced with longer sequences, validating superior efficiency of HiFi-Mesh in long sequence processing tasks.
\section{5\ \ \ \ Conclusion}
This paper proposes HiFi-Mesh, a method for rapidly generating detail-rich 3D meshes. The approach leverages compact coherent structural representations and local detail information in latent spaces as autoregressive dependencies, effectively reducing reliance on long historical token sequences and significantly lowering computational overhead. This improvement enables longer maximum generatable sequences, achieving higher structural details. Additionally, we introduce the AdaGraph strategy, which decouples the generation process spatiotemporally into multiple parallel sub-sequences, substantially improving inference speed.
Experimental results demonstrate that HiFi-Mesh achieves superior performance in generation speed, structural details, and geometric consistency with input point clouds. This provides an insightful solution to long sequence generation problems and marks an important step toward automatically generating artist-friendly 3D meshes.
\section{Acknowledgements}
This work was supported by the grant from Science and Technology Development Fund of Macao (0009/2025/ITP1). This work is supported by Macao Polytechnic University Grant (RP/FCA-08/2024). The work was supported in part by the National Natural Science Foundation of China under Grant 62301310, and in part by Sichuan Science and Technology Program under Grant 2024NSFSC1426.

\bibliography{aaai2026}

\end{document}